\documentclass[runningheads]{llncs}

\usepackage{eccv}

\usepackage{eccvabbrv}

\usepackage{graphicx}
\usepackage{booktabs}

\usepackage[accsupp]{axessibility}  

\usepackage{hyperref}

\usepackage{orcidlink}

\begin{document}

\title{AvatarPose: Avatar-guided 3D Pose Estimation \\
of Close Human Interaction \\
from Sparse Multi-view Videos} 

\titlerunning{AvatarPose}

\def\thefootnote{*}\footnotetext{Equal contribution}
\def\thefootnote{\dag}\footnotetext{Now at HKUST(GZ)$\&$HKUST}

\author{Feichi Lu\inst{1}$^*$\orcidlink{0009-0007-4017-9606} \and
Zijian Dong\inst{1,2}$^*$\orcidlink{0009-0003-4150-5072} \and
Jie Song\inst{1}$^\dag$\orcidlink{0009-0003-7484-1937} \and
Otmar Hilliges\inst{1}\orcidlink{0000-0002-5068-3474}} 

\authorrunning{F. Lu et al.}

\institute{Department of Computer Science, ETH Zürich, Switzerland, \and Max Planck Institute for Intelligent Systems, Germany
}

\maketitle

\begin{center}
    \centering
    \captionsetup{type=figure}
    \includegraphics[width=\textwidth]{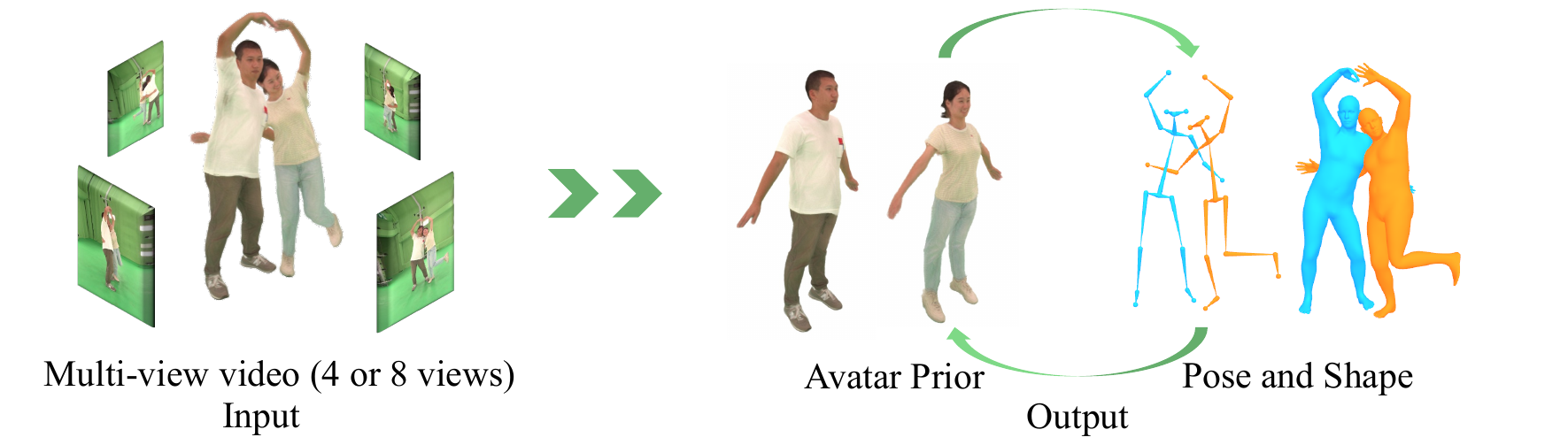}
    \captionof{figure}{We propose AvatarPose, a method for estimating the 3D poses and shapes of multiple closely interacting people from multi-view videos. To this end, we first reconstruct the avatar of each individual and leverage the learned personalized avatars as priors to refine poses via color and silhouette rendering loss. We alternate between avatar refinement and pose optimization to obtain the final pose estimation.  }
    \label{fig:teaser}
\end{center}

\begin{abstract}

 Despite progress in human motion capture, existing multi-view methods often face challenges in estimating the 3D pose and shape of multiple closely interacting people. This difficulty arises from reliance on accurate 2D joint estimations, which are hard to obtain due to occlusions and body contact when people are in close interaction. To address this, we propose a novel method leveraging the personalized implicit neural avatar of each individual as a prior, which significantly improves the robustness and precision of this challenging pose estimation task. Concretely, the avatars are efficiently reconstructed via layered volume rendering from sparse multi-view videos. The reconstructed avatar prior allows for the direct optimization of 3D poses based on color and silhouette rendering loss, bypassing the issues associated with noisy 2D detections. To handle interpenetration, we propose a collision loss on the overlapping shape regions of avatars to add penetration constraints. Moreover, both 3D poses and avatars are optimized in an alternating manner. Our experimental results demonstrate state-of-the-art performance on several public datasets.
 

\end{abstract}    

\section{Introduction}




People frequently engage in interactions with each other in daily life to offer physical support or convey emotions. For AI systems to interpret 3D human interactions, a foundational step is to reconstruct 3D human poses and shapes using consumer-grade sensors, such as cameras. 
However, in closely interacting scenarios, such as hugging or kissing, 3D pose estimations face challenges due to strong occlusions. The problem of depth ambiguity worsens when close contact is involved, making it more difficult to predict 3D poses from 2D images or videos. Consequently, a multi-camera setup becomes essential to provide additional observations and to address depth ambiguity in pose estimation. 


Despite the ubiquity of close human interactions, the study of estimating such human motions is scarce. Most previous multi-human methods with a multi-view setup~\cite{belagiannis1,belagiannis2,ershadi,mvpose,4Dassociation, voxelpose,faster_voxelpose,panoptic,shelf} focus on scenarios where people are at a distance from each other. 
Some methods~\cite{belagiannis1,belagiannis2,ershadi,mvpose,4Dassociation} typically formulate the problem as
a cross-view matching problem, relying heavily on 2D joint estimations for subsequent 3D triangulation. These methods demonstrate high sensitivity to noisy or missing 2D joint estimations, particularly when occlusion occurs. Another group of learning-based methods~\cite{voxelpose,faster_voxelpose,direct_regression,graph-based} attempts to integrate 2D features from each view into a 3D voxel space and predict 3D human poses from identified 3D subvolumes of each individual. These methods are more robust to occlusion, but they encounter challenges with generalization and are sensitive to distribution shifts due to the lack of annotated 3D data. 
To correct abnormal or missing pose estimations, some methods~\cite{shape_aware, mvpose2021} leverage parametric body models like SMPL~\cite{smpl} as full-body priors and fit these 3D models to 2D joint estimations. Although this approach alleviates the issue of abnormal or missing joint estimations, it remains constrained by noisy 2D detections. 

Embracing the challenging problem of pose estimation for close interactions, we propose a novel method to estimate the 3D poses and shapes of multiple people observed from a sparse set of cameras. Our goal is to ensure that even in close contact, the estimated human poses and shapes are accurate and free from interpenetrations. The key to our method is reconstructing implicit textured avatars of each individual in the scene and leveraging them as a strong personalized prior for pose optimization (\Cref{fig:teaser}).  In contrast to relying solely on noisy 2D joint detections, this textured avatar prior enables us to leverage pixel-wise color and silhouette information for pose refinement, significantly increasing the robustness and accuracy of our method (\Cref{tab:ablation}). Meanwhile, the reconstructed avatar provides crucial geometric and appearance information to avoid collisions between individuals. Compared to methods using SMPL body shape~\cite{selfcontact, chi3d} to penalize collisions, our implicit avatar model contains a more detailed clothed shape and additional appearance cues and enables efficient computation of penetration loss. Due to the mutually beneficial relationship between avatar and pose, we alternate between pose optimization and avatar refinement. 



More specifically, to accelerate the learning of avatars, we model each human individual in canonical space using an efficient neural radiance field variant in Instant NGP~\cite{instantngp} and combine it with an efficient SMPL-based deformation module~\cite{smpl}. To learn and render avatar models of multiple people, we adapt layered volume rendering~\cite{stnerf, easymocap} to our avatar model. This adaptation allows us to jointly optimize all avatar models through a straightforward rendering loss. 
Once learned, the avatar can be animated and rendered based on pose parameters at interactive rates, thus naturally serving as an efficient personalized textured prior for pose optimization. With the learned personalized prior, we optimize pose via a novel objective function. Different from previous methods based on 2D reprojection error of joints, we directly optimize pose parameters via minimizing color and silhouette rendering losses while keeping the learned avatar model fixed. To prevent interpenetration between human individuals, a collision loss is introduced by penalizing the situation when a 3D point is occupied by multiple avatars. To remove artifacts in the initial avatar due to imperfect pose initialization, we further refine avatars based on optimized poses. Throughout the optimization process, both initial personalized avatar models and SMPL parameters are optimized in an alternating manner, motivated by the insight that accurate 3D pose estimations improve avatar learning, and improved avatar models, in turn, increase the precision of overall pose estimations.



We experimentally demonstrate that our method significantly outperforms previous state-of-the-art methods on several public datasets both quantitatively (\Cref{tab:comparison_hi4d} and \Cref{tab:comparison_chi3d}) and qualitatively (\Cref{fig:comparison}) especially when people are in close interaction. 
In summary, our contributions are:
\begin{itemize}
 \item We propose a pipeline that efficiently creates implicit neural avatars of closely interacting people and leverages the learned avatars as priors to optimize poses. 
 
 \item The avatar prior enables us to design a novel objective function that leverages color and mask rendering loss for pose optimization. We show the superiority of this loss function compared to the 2D reprojection error of 3D joints used by most of the previous methods. 
 
 \item Based on the learned avatar, a collision loss is introduced to avoid penetration when individuals are in contact.

 
\end{itemize}

\section{Related Work}

\subsection{Multi-Person 3D Pose Estimation} \label{sec:multi-person}

 Despite significant progress in multi-human 2D pose estimation~\cite{openpose, deepcut, higherHRnet, rmpe, cascaded, tokenpose} and 3D pose estimation from monocular images or video~\cite{smplify, mono_1, mono_2, mono_3, mono_4, PARE, ROMP, BEV,pymaf,spec, agora, hmor, single_scene, rebuttal_1}, the reconstruction accuracy is still limited due to depth ambiguity and strong occlusions when humans are in close contact. A multi-view setting~\cite{seg_motion, belagiannis2, ershadi, mvpose, 4Dassociation, closemocap, faster_voxelpose, direct_regression, plane-sweep, pandanet, regress_1, graph-based} helps to alleviate these challenges. One straightforward idea of most methods~\cite{belagiannis1, belagiannis2, ershadi, mvpose,4Dassociation,quickpose} is to formulate the problem into cross-view matching and association problems. MVPose~\cite{mvpose} performs 2D person parsing in each image and leverages cross-view person matching to infer 3D pose. 4DAssociation~\cite{4Dassociation} additionally adds tracking into this process to form a unified graph for associating 4D information.  However, these methods are sensitive to the noisy estimation of 2D pose. In contrast to these matching-based methods, some recent methods~\cite{voxelpose,faster_voxelpose,voxeltrack, heatmap,tempo, direct_regression,tessetrack,graph-based,plane-sweep,vtp,pandanet,regress_1} directly learn deep neural networks to regress poses. Faster Voxelpose~\cite{faster_voxelpose} employs the feature volume proposed by Voxelpose~\cite{voxelpose} and enhances computational efficiency. Graph~\cite{graph-based} designs three graph neural network models for human center detection and pose estimation. MvP~\cite{direct_regression} simplifies the multi-person pose estimation by direct regression using the transformer model. A concurrent method CloseMocap~\cite{closemocap} proposes to learn a model from a synthetic dataset simulating occlusion situations. However, heavily relying on the 2D or 3D features as input during training, these learning-based methods suffer from generalization issues when subjects, motions, and camera configurations change.

To further improve the robustness, a statistical parametric body model such as SMPL~\cite{smpl} is explored in ~\cite{shape_aware} as a regularization prior for 3D joint refinement. Some follow-ups~\cite{lightweight, mvpose2021} show that parametric models help in correcting implausible 3D pose estimates and filling in missing joints. However, this coarse body prior highly relies on aligning the 3D joints to 2D pose estimations, which are inaccurate when occlusion happens.  Different from all of these previous methods, we explore the usage of personalized textured avatar models as priors to refine human poses. This prior enables us to leverage color and silhouette information from multi-view observation for refining poses of closely interacting humans.


\subsection{3D Human Modeling}

Parametric human body models~\cite{smpl,scape,statistical,star,ghum} can represent minimally clothed human shapes by deforming a template mesh.  It is challenging to extend this explicit representation for modeling clothed humans due to the fixed topology and
resolution. To overcome this limitation, methods such as SNARF~\cite{snarf} and SCANimate~\cite{scanimate} propose to model articulated human shapes based on 3D implicit representations. Many works~\cite{neuralbody, arah, te_neural, v2a, instantavatar, humannerf, animnerf, pina,ipnet,dsfn} fit implicit neural fields to RGB or RGB-D videos by neural rendering to reconstruct the shape and appearance of a single human body. However, when applied to a multi-human scene, these methods are not able to achieve good fidelity due to strong occlusions. Recent methods including ST-NeRF~\cite{stnerf} and ~\cite{easymocap} leverage layered neural representation to model multiple humans with sparse multi-view videos and thus can generate novel view synthesis of dynamic multiple humans. The main problem of all aforementioned approaches, however, is their high reliance on 3D human pose estimation~\cite{deepmulticap, stnerf}: the deformation of the human model requires accurate human poses, which is hard to obtain when people are in close interaction. To address this challenge, our method is orthogonal to others, aiming at leveraging the learned avatar as priors for pose estimation in close interaction.

\subsection{3D Human Datasets for Close Interaction}

Most of the existing datasets~\cite{shelf, panoptic, dataset_1} like Shelf and Panoptic studying multi-person pose estimation focus on the scene where people are at a distance from each other and rarely involved in close interactions. To study close interactions among people, MultiHuman~\cite{deepmulticap} dataset captures multi-person interaction with some close interactions and occlusions. ExPI~\cite{expi} creates a multi-person extreme motion dataset with close interactions, but it focuses mainly on motion prediction for future frames instead of pose estimation from sparse views. CHI3D~\cite{chi3d} captures two-person interaction datasets and proposes to learn a contact estimation module from annotations to improve the precision of pose estimation.  The most recent work Hi4D~\cite{Hi4D} creates a challenging dataset of physically close human interaction and proposes a method to disentangle human bodies and estimate poses. However, this method relies on the 3D ground truth of clothed human meshes captured with expensive 3D body scanners. 
Due to the reliance on the limited annotated 3D data, this method faces challenges in generalization with different people and camera configurations. 
In contrast to all the works before, we intend to solve the problem of pose estimation in close human interaction without requiring accurate 3D scans or other training data.

\section{Method}

\begin{figure*}[h]
  \centering
   \includegraphics[width=1\linewidth]{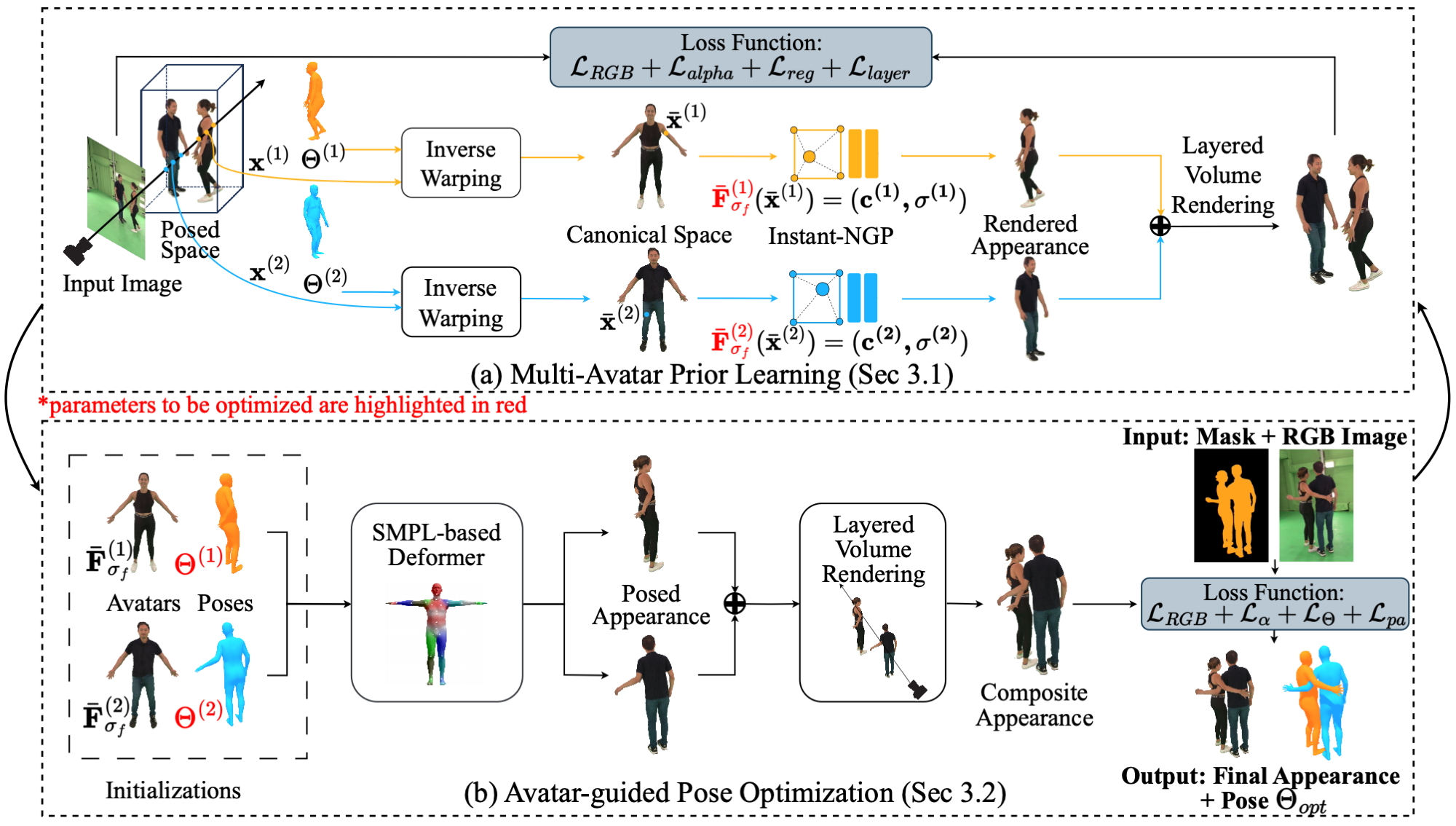}
   \caption{\textbf{Method Overview}: Our method consists of two modules:   \textit{(a) Multi-Avatar Prior Learning:} Given the input multi-view images and estimated poses $\mathbf{\Theta}^{(l)}$, we sample points $\mathbf{x}^{(l)}$ for each individual $l$ along the rays in posed space and warp these points into canonical space and calculate their color $\mathbf{c}^{(l)}$ and density $\mathbf{\sigma}^{(l)}$ via the canonical appearance network $\mathbf{\bar{F}}_{\sigma_{f}}^{(l)}$. We leverage layered volume rendering~\cite{stnerf} to attain the final pixel color and compare it with the original input image to optimize the parameters of avatars. 
   \textit{(b) Avatar-guided Pose Optimization:} Given learned avatar model $\mathbf{\bar{F}}_{\sigma_{f}}^{(l)}$ and initial poses $\mathbf{\Theta}^{(l)}$ of each individual $l$, we deform all of the avatars based on SMPL-based deformer and render them jointly via layered volume rendering. We compare the composite rendering with input observation and minimize the RGB and mask rendering loss to optimize poses. A collision loss is additionally introduced to avoid interpenetration. Finally, we alternate between two modules to obtain the final result. For clarity, the parameters to be optimized are marked as red in each module. }
   \label{fig:pipeline}
\end{figure*}

Given a dynamic scene captured by a sparse set of RGB cameras, our goal is to estimate the 3D pose and shape of multiple people even if they interact closely. To address this challenging task, our key idea is to first reconstruct the personalized avatar of each individual in the scene and leverage them as a strong prior to refine the appearance and pose in an alternating manner. An overview of our method is shown in \Cref{fig:pipeline}.

We first introduce an efficient pipeline to create avatars of multiple people in a scene (\Cref{sec:avatar_learning} and \Cref{fig:pipeline}(a)). Specifically, we leverage an accelerated neural radiance field to represent the shape and appearance of each individual in canonical space and deform it at an interactive rate. We then adapt layered volume rendering to our pipeline, which composites the rendering of avatars into one image, thus enabling direct learning from multi-view video inputs.

Thanks to the learned avatar prior for each individual, we can enhance 3D pose optimization via a combination of RGB and silhouette rendering loss (\Cref{sec:pose_opt} and \Cref{fig:pipeline}(b)). While previous work heavily relies on noisy 2D joint detection, we show that employing such pixel-wise color and silhouette information can largely increase precision and robustness. Moreover, a collision loss is introduced to avoid interpenetration. Finally, we alternate between avatar learning and pose optimization to get complete and accurate 3D human poses.

\subsection{Multi-Avatar Prior Learning}
\label{sec:avatar_learning}

\paragraph{Avatar Model}

 We represent each human individual in canonical space using an accelerated neural radiance field~\cite{instantngp} and model shape-aware articulated deformation based on SMPL~\cite{smpl}.
\begin{itemize}

    \item \textbf{Canonical Appearance Representation:} To model human shape and appearance, we create canonical radiance field $\mathbf{\bar{F}}_{\sigma_{f}}^{(l)}$ for each human instance $l \in [1, L]$, where $L$ is the number of humans in the scene. $\mathbf{\bar{F}}_{\sigma_{f}}^{(l)}$ takes a 3D point $\bar{\mathbf{x}}^{(l)}$ as input and predicts its density $\sigma^{(l)}$ and color $\mathbf{c}^{(l)}$. Following Instant-NGP~\cite{instantngp} and InstantAvatar~\cite{instantavatar} to accelerate the rendering, $\mathbf{\bar{F}}_{\sigma_{f}}^{(l)}$ is parameterized via using a hash
    table to store feature grids at different scales.
  
    \item {\textbf{Pose Representation}}: We represent the 3D pose and underlying body shape for all human instances by SMPL parameters $\mathbf{\Theta}=\{\mathbf{\Theta}^{(l)}\}_{l\in[1, L]}$. For each human $l$, $\mathbf{\Theta}^{(l)} = \{\beta^{(l)}, \theta^{(l)}, t^{(l)}\}$ contains shape parameters $\beta^{(l)} \in \mathbb{R}^{10}$, pose parameters $\theta^{(l)} \in \mathbb{R}^{72}$ and translation $t^{(l)} \in \mathbb{R}^{3}$ of SMPL. 

    \item \textbf{Deformer:} To enable animation given targeted poses $\mathbf{\Theta}^{(l)}$, we require the radiance field in the posed
    space.  Given a point $\mathbf{x}^{(l)}$ in deformed space of human $l$, we determine the corresponding canonical point $\bar{\mathbf{x}}^{(l)}$ by inverse linear blend skinning(LBS)~\cite{smpl}: $\bar{\mathbf{x}}^{(l)}(\mathbf{x}^{(l)}, \mathbf{\Theta}^{(l)}) = (\sum_{i=1}^{n_{b}}w_{i}(\mathbf{\Theta}^{(l)})\mathbf{B}_{i} (\mathbf{\Theta}^{(l)}))^{-1}\mathbf{x}^{(l)}$, where $\mathbf{B}_{i}$ is the rigid bone transformation matrix for joint $i \in \{1,...,n_b\}$ under pose $\mathbf{\Theta}^{(l)}$. $w_{i}$ is the skinning weights of the nearest neighbor of $\mathbf{x}^{(l)}$ in the deformed SMPL vertices. We obtain the radiance field at the point $\mathbf{x}^{(l)}$ by evaluating the canonical appearance field at the corresponding point $\bar{\mathbf{x}}^{(l)}$. 
\end{itemize}

\paragraph{Layered Volume Rendering}
To obtain the pixel value for a ray $\mathbf{r} \in \mathcal{R}$, we raycast every human instance separately with a layered rendering strategy similar to ST-NeRF~\cite{stnerf}. Specifically, we first calculate the intersection points between the ray and the 3D bounding box of each human instance and uniformly sample points in each bounding box. To distinguish different identities, we assign each sampled point $\mathbf{x}_{i}$ a one-hot representation $\mathbf{m}_{i} = [m_{i}^{(1)}, \cdots, m_{i}^{(L)}]$ to indicate which human identity it belongs to. After sorting all sampled points by their depth values and calculating their corresponding color $\mathbf{c}_{i}$ and density $\sigma_i$ from the avatar model, if $m_{i}^{(l)} = 1$, we compute 
\begin{equation} \label{eq:color}
    \mathbf{c}_{i}, \sigma_{i}  = \mathbf{\bar{F}}^{(l)}_{\sigma_{f}}(\mathbf{\bar{x}}_{i}(\mathbf{x}_{i}, \mathbf{\Theta}^{(l)})).
\end{equation}

The color of each ray is computed via numerical integration~\cite{nerf}.
\begin{equation}\label{eq:nerf}
        \hat{\mathbf{C}}(\mathbf{r})  = \sum_{i=1}^{N}\alpha_{i}\prod_{j < i} (1-\alpha_{j})\mathbf{c}_{i} \quad
        \alpha_{i}  = 1 - \exp(-\sigma_{i}\delta_{i}), 
\end{equation}

where $\delta_i$ is the distance between samples. The accumulated alpha value, which represents ray opacity, can be computed via:
\begin{equation} \label{eq:silhouette}
      \alpha (\mathbf{r}) = \sum_{i=1}^{N}\alpha_{i}\prod_{j < i}(1-\alpha_{j}). \\
\end{equation}

For each human identity $l\in[1, L]$, the corresponding instance ray opacity can be calculated via:
\begin{equation}
      \alpha^{(l)} (\mathbf{r}) = \sum_{i=1}^{N}\alpha_{i}\prod_{j < i}(1-\alpha_{j}) m_{i}^{(l)}.
\end{equation}



\paragraph{Training}
The overall training process is shown in \Cref{fig:pipeline}(a). For training avatar layers, we minimize the Huber loss $\rho$ between the predicted pixel color $\hat{\mathbf{C}}(\mathbf{r})$ and the ground truth pixel color $\mathbf{C}_{gt}(\mathbf{r})$:
\begin{equation}
    \mathcal{L}_{RGB} = \frac{1}{\mid \mathcal{R}\mid}\sum_{r\in\mathcal{R}}\rho(\lVert \hat{\mathbf{C}}(\mathbf{r}) - \mathbf{C}_{gt}(\mathbf{r}) \rVert).
\end{equation}


Since instance segmentation of human performers is hard to obtain and is not accurate, we choose foreground segmentation as our mask supervision, which is obtained via SAM-Track~\cite{trackingsam}. We apply a loss for optimizing the rendered alpha values $\alpha$ to reduce the artifacts in the floating area:

\begin{equation}
    \mathcal{L}_{alpha} =\frac{1}{\mid \mathcal{R}\mid}\sum_{\mathbf{r} \in \mathcal{R}}(\alpha(\mathbf{r}) - \alpha_{SAM}(\mathbf{r}))^{2}.
\end{equation}

Following~\cite{easymocap}, we add a regularization loss for instance alpha values to make sure every pixel can only be rendered from one human layer:

\begin{equation}
    \mathcal{L}_{layer} = - \frac{1}{L\mid \mathcal{R} \mid}\sum_{\mathbf{r}\in \mathcal{R}}\sum_{l=1}^{L}\alpha^{(l)}(\mathbf{r})\log(\alpha^{(l)}(\mathbf{r})).
\end{equation}

Similar to~\cite{instantavatar}, we also add hard surface and density regularization terms in the learning process. More training details can be found in the Supp Mat.

\subsection{Avatar-guided Pose Optimization } \label{sec:pose_opt}

Equipped with learned avatars obtained in \Cref{sec:avatar_learning}, we aim to estimate the 3D shape and pose of multiple humans with close physical contact. To achieve this, we leverage the avatars as priors to handle the challenging pose ambiguities caused by contact. Specifically, we first initialize pose parameters using an off-the-shelf 3D pose estimator~\cite{4Dassociation} and refine the pose via a rendering loss between rendered posed avatars and 2D observations (\Cref{fig:pipeline}(b)). Since the initial imperfect pose estimations may cause artifacts in avatar reconstruction, we further refine the weights of the avatar model using the optimized pose. This process is formulated as an alternating optimization to refine both poses and avatars. 





\paragraph{Initialization} \label{sec:pose_init}

The initial 3D pose proposals are estimated by leveraging the off-the-shelf 3D human pose estimator~\cite{4Dassociation}. After that, we register a SMPL model to the estimated 3D joints to obtain the initial pose parameters $\mathbf{\Theta_{0}}$. Given these estimated poses, the initial avatar model of each individual is further learned from multi-view videos.



\paragraph{Objective}

To leverage avatars as priors to tackle challenges caused by contact, we optimize the following objective:
\begin{equation}
\begin{split}
    \mathcal{L}(\mathbf{\Theta}) & = \lambda_{RGB}\mathcal{L}_{RGB}(\mathbf{\Theta}) + \lambda_{\alpha}\mathcal{L}_{\alpha} (\mathbf{\Theta})  \\
    &+\lambda_{reg}\mathcal{L}_{reg} (\mathbf{\Theta}
    ) + \lambda_{pa}\mathcal{L}_{pa} (\mathbf{\Theta}).
\end{split}
\end{equation}

Here, $\mathbf{\Theta}$ is the SMPL parameters to be optimized for all avatars, which is also consistent with the pose parameters to deform the avatar model (\Cref{sec:avatar_learning}). Following~\cite{humannerf}, we represent  $\mathbf{\Theta}$ as $\mathbf{\Theta_0} + \text{MLP}(\mathbf{\Theta_{0}})$ and refine poses by changing the parameters of the neural network. This representation empirically shows more robust results compared to directly optimizing SMPL parameters.

Different from previous methods~\cite{shape_aware, 4Dassociation, mvpose2021,voxelpose}, the personalized multi-avatar prior allows us to leverage appearance and silhouette information to refine initial poses. 
Specifically, we calculate $\mathcal{L}_{RGB}$ to ensure the color consistency between the rendered pixel $\hat{\mathbf{C}}(\mathbf{r},\mathbf{\Theta})$ (\Cref{eq:color} and \Cref{eq:nerf}) of deformed avatar with poses $\mathbf{\Theta}$ and the corresponding ground-truth pixel color $\mathbf{C}_{gt}(\mathbf{r})$. 


\begin{equation}
    \mathcal{L}_{RGB} (\mathbf{\Theta}) = \frac{1}{\mid \mathcal{R} \mid}\sum_{r\in\mathcal{R}}\rho(\lVert \hat{\mathbf{C}}(\mathbf{\mathbf{\mathbf{r}, \Theta}}) - \mathbf{C}_{gt}(\mathbf{r}) \rVert).
\end{equation}


Additionally, a cross-entropy loss $\mathcal{L}_{\alpha}$ is introduced to ensure that the rendered mask $\alpha(\mathbf{r}, \mathbf{\Theta})$ (\Cref{eq:silhouette}) of the reposed avatar is aligned with the estimated SAM-Track mask $\alpha_{SAM}(\mathbf{r})$ by:

\begin{equation}
    \mathcal{L}_{\alpha} (\mathbf{\Theta}) = -\sum_{\mathbf{r}\in\mathcal{R}}\alpha_{SAM}(\mathbf{r})\log(\alpha( \mathbf{r}, \mathbf{\Theta}))
\end{equation}

To penalize the unnatural poses and avoid elbows and knees bending in the wrong direction, we add an L2 regularization term and combine it with the pose prior in SMPLify~\cite{smplify} to constrain physically implausible joint rotation:
\begin{equation}
    \mathcal{L}_{reg}(\mathbf{\Theta}) = \lVert\mathbf{\Theta}\rVert_{2} + \lambda \sum_{i\in I}\exp(\mathbf{\Theta}_i),
\end{equation}

where $I$ is the set of pose indices corresponding to elbows and knees.

%
A key challenge to correctly estimate poses in close interaction is to handle interpenetration. Since every avatar is modeled separately, the surfaces tend to intersect when they are in contact. To handle this, we first select sampled points inside multiple instances as $\mathcal{S} = \{\mathbf{x}_{i} \mid \alpha_{i}^{(p)}>0, \alpha_{i}^{(q)}>0, p, q\in [1, L], p\neq q\}$ ($\alpha_{i}^{(p)}$ is calculated from \Cref{eq:color} and \Cref{eq:nerf} corresponding with a point $\mathbf{x}_{i}$ with $m_{i}^{(p)} = 1$). We then propose a collision loss $\mathcal{L}_{pa}$ for penalizing penetration: 

\begin{equation}\label{eq:pene_avatar}
    \mathcal{L}_{pa} (\mathbf{\Theta}) = \frac{1}{\mid \mathcal{S}\mid}\sum_{\mathbf{x}_i \in \mathcal{S}}\alpha_{i}^{(p)} (\mathbf{\Theta})\alpha_{i}^{(q)} (\mathbf{\Theta})
\end{equation}

Intuitively, this loss guarantees every sample point in 3D space can not be occupied by multiple avatar models simultaneously, which guarantees better pose estimation in close contact.

\paragraph{Alternating Optimization}

Since artifacts sometimes appear on initial avatars due to imperfect pose initialization, we further refine avatars based on optimized poses via minimizing the loss function in \Cref{sec:avatar_learning}. Finally, the optimization of poses and avatars is formulated in an alternating fashion for $N$ steps. More details of optimization can be found in the Supp Mat.










\begin{figure*}[h]

   \includegraphics[width=\textwidth]{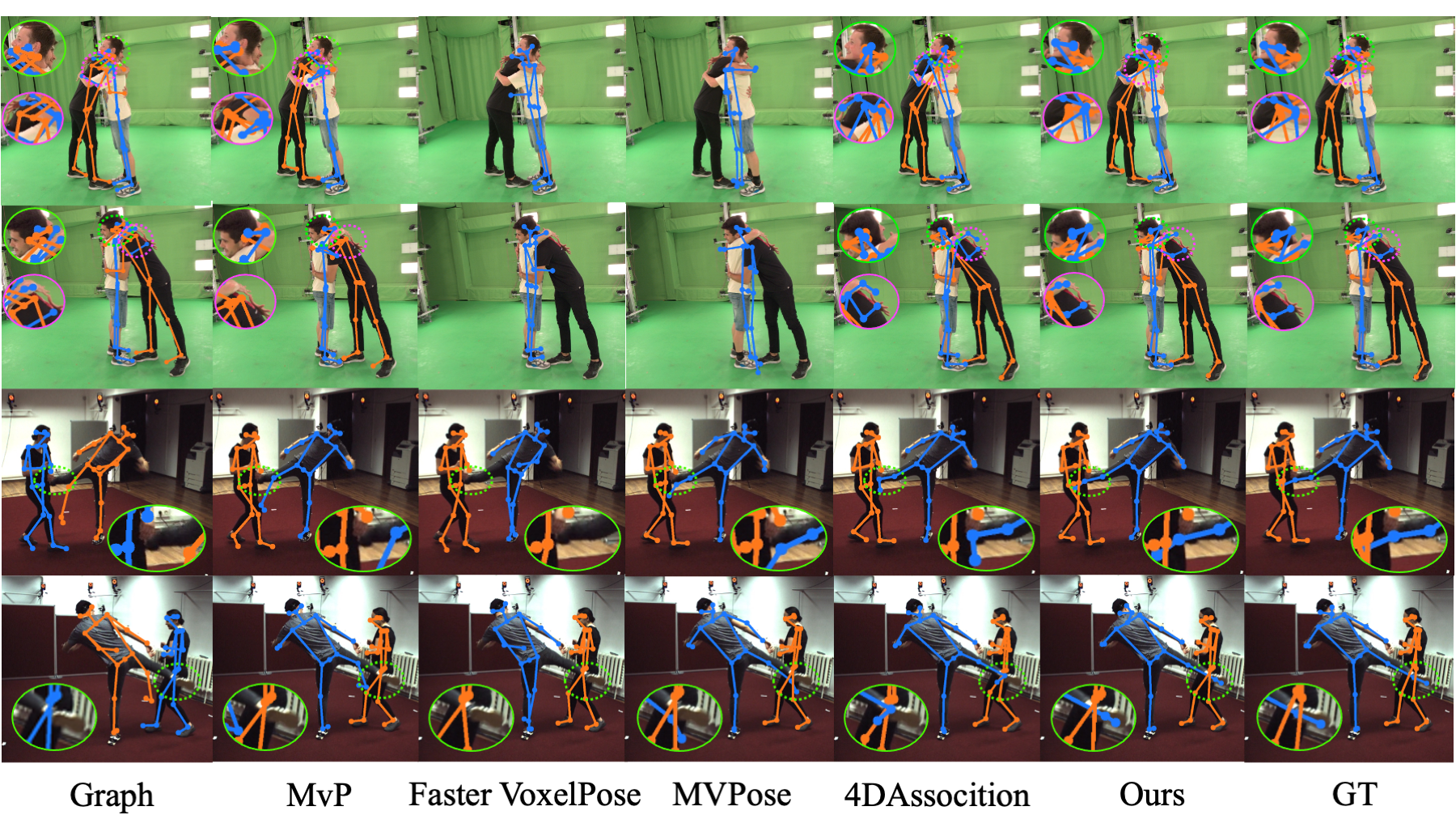}
   \caption{\textbf{Qualitative Comparison with SotA methods~\cite{mvpose, 4Dassociation, faster_voxelpose, direct_regression, graph} on Hi4D and CHI3D.} We show two examples from the Hi4D and CHI3D datasets compared with Graph, MvP, Faster VoxelPose, MVPose, and 4DAssociation. For each example, we show 2D projections on two sampled views.}
   \label{fig:comparison}
\end{figure*}

\section{Experiments}
\paragraph{Datasets.}
We mainly evaluate our proposed method on Hi4D~\cite{Hi4D} and CHI3D~\cite{chi3d}, which are challenging datasets of two humans in close interaction. To demonstrate the generalization ability of our method for more than two people, we also evaluate on Shelf~\cite{shelf} and MultiHuman~\cite{deepmulticap} including three or four people. More details are shown in the Supp Mat.


\paragraph{Metrics.}
We use the Mean Per Joint Position Error (MPJPE)~\cite{voxelpose} to measure the distance between the ground truth 3D poses and the estimated poses. We also choose the Percentage of Correct Parts (PCP3D) metric~\cite{mvpose} to calculate the percentage of correct parts. $\text{AP}_{K}$~\cite{voxelpose} and Recall~\cite{shape_aware} are also leveraged to evaluate performance. More details are shown in the Supp Mat.




\paragraph{Baselines.}
We compare our method on the Hi4D dataset with state-of-the-art methods in three categories discussed in \Cref{sec:multi-person}. For learning-based methods, we choose Graph~\cite{graph}, MvP~\cite{direct_regression}, and Faster VoxelPose~\cite{faster_voxelpose} and fine-tune these models on a subset of the evaluating datasets. For pure association-based methods, we choose 4DAssociation~\cite{4Dassociation} and MVPose*~\cite{mvpose}. MVPose~\cite{mvpose2021} adds temporal tracking and SMPL prior to MVPose*~\cite{mvpose} and is regarded as a SMPL-guided method. More details about baselines can be found in the Supp Mat.



\begin{table*}[h]
  \centering
  \begin{tabular}{@{}lcccccccccc@{}}
    \toprule
    \textbf{Method} & \textbf{MPJPE}(mm) $\downarrow$ & \textbf{PCP}(\%) $\uparrow$  & $\textbf{AP}_{50}$ $\uparrow$ & $\textbf{AP}_{100}$ $\uparrow$ & \textbf{Recall}(\%) $\uparrow$ \\
    \midrule
    MvP~\cite{direct_regression} & 92.77 & 74.14  & 41.59 & 63.86 & 93.84 \\
    Graph~\cite{graph-based} & 89.62 & 71.55  & 44.75 & 67.33 & 93.31 \\
    Faster VoxelPose~\cite{faster_voxelpose} & 68.40 & 73.67 & 44.05 & 68.70 & 83.55\\
    MVPose*~\cite{mvpose} & 53.05 & 87.57 & 67.97 & 80.28 & 93.80\\
    MVPose~\cite{mvpose2021} & 42.63 & 90.76  & 71.79 & 90.19 & 93.30\\
    4DAssociation~\cite{4Dassociation} & 41.29 & 88.62  & 80.87 & 97.27 & \textbf{98.78}\\
    \midrule
    Ours & \textbf{32.10} & \textbf{96.90}  & \textbf{91.48} & \textbf{97.33} & \textbf{98.78} \\
    \bottomrule
  \end{tabular}
  \caption{\textbf{Quantitative Comparison with SotA on the Hi4D ~\cite{Hi4D} Dataset (8 views).} We compare our method with MvP~\cite{direct_regression}, Graph~\cite{graph-based}, Faster VoxelPose~\cite{faster_voxelpose}, MVPose*~\cite{mvpose}, MVPose~\cite{mvpose2021} and 4DAssociation~\cite{4Dassociation}. We report MPJPE, PCP, $\text{AP}_{K}$, and Recall metric for all methods. }
  \label{tab:comparison_hi4d}
\end{table*}

\begin{table*}[h]
  \centering
  \resizebox{\columnwidth}{!}{
  \begin{tabular}{@{}lcccccccccc@{}}
    \toprule
    \textbf{Method} & \textbf{MPJPE}(mm) $\downarrow$ & \textbf{PCP}(\%) $\uparrow$  & $\textbf{AP}_{50}$ $\uparrow$ & $\textbf{AP}_{100}$ $\uparrow$ & \textbf{Recall}(\%) $\uparrow$ \\
    \midrule
    MvP~\cite{direct_regression} & 55.38 & 89.47 & 63.58 & 92.53 & 99.06 \\
    Graph~\cite{graph-based} & 45.33 & 92.46 & 74.02 & 95.25 & 99.17 \\
    Faster VoxelPose~\cite{faster_voxelpose} & 67.81 & 78.41 & 29.28 & 82.88  & 93.34\\
    MVPose*~\cite{mvpose} & 50.42 & 90.39 & 69.13 & 75.72 & 88.72\\
    MVPose~\cite{mvpose2021} & 34.05 & 93.35  & 79.94 & 86.91 & 88.18\\
    4DAssociation~\cite{4Dassociation} & 37.47 & 99.30 & 89.66 & 98.67 & \textbf{99.85}\\
    \midrule
    Ours & \textbf{32.98} & \textbf{99.79} & \textbf{93.20} & \textbf{99.79} & \textbf{99.85} \\
    \bottomrule
  \end{tabular}}
  \caption{\textbf{Quantitative Comparison with SotA on the CHI3D ~\cite{chi3d} Dataset (4 views).} We compare our method with Faster VoxelPose~\cite{faster_voxelpose}, MVPose*~\cite{mvpose}, MVPose~\cite{mvpose2021} and 4DAssociation~\cite{4Dassociation}. We report MPJPE, PCP, $\text{AP}_{K}$, and Recall metric for all methods. }
  \label{tab:comparison_chi3d}
\end{table*}

\subsection{Comparison to SotA}

\Cref{tab:comparison_hi4d} and \Cref{tab:comparison_chi3d} summarizes our quantitative comparisons on Hi4D and CHI3D with SotA (State-of-the-art) methods. Our method largely outperforms the other SotA methods in all of the metrics including MPJPE, PCP3D, and $\text{AP}_{K}$.  More comparisons on Shelf and MultiHuman are shown in the Supp Mat.

\paragraph{Comparison with Graph~\cite{graph-based}, MvP~\cite{direct_regression}, Faster VoxelPose~\cite{faster_voxelpose}.} Our method outperforms Graph, MvP, and Faster VoxelPose on Hi4D and CHI3D. These methods are prone to overfitting to training pose distributions, thus struggling with challenging poses of interacting actors. Specifically, Faster VoxelPose sometimes misses actors in close contact, leading to a relatively low recall. While Graph and MvP have better recall, Graph fails to consistently track actors across frames, and MvP results in many misaligned joints between actors.


\paragraph{Comparison with MVPose~\cite{mvpose2021}.} Our method achieves much better MPJPE and precision than MVPose. MVPose relies heavily on noisy 2D joint detected in close interactions. In contrast, our method leverages color and silhouette rendering loss to optimize poses, leading to robustness to occlusions. Additionally, this top-down method cannot detect actors correctly with strong occlusions, as shown in \Cref{fig:comparison}.


\paragraph{Comparison with 4DAssociation~\cite{4Dassociation}.}
Finally, we compare our method with the bottom-up association method. As shown in \Cref{tab:comparison_hi4d} and \Cref{tab:comparison_chi3d}, our method outperforms 4DAssociation in most metrics. 
\Cref{fig:comparison} shows that when actors are close, this bottom-up method is inclined to associate joints with the wrong human instances. This is because they solve the joint association with a greedy algorithm, which is sensitive to missing and inaccurate 2D joint detections.  

\begin{figure*}[h]
    \centering
    \includegraphics[width=\linewidth]{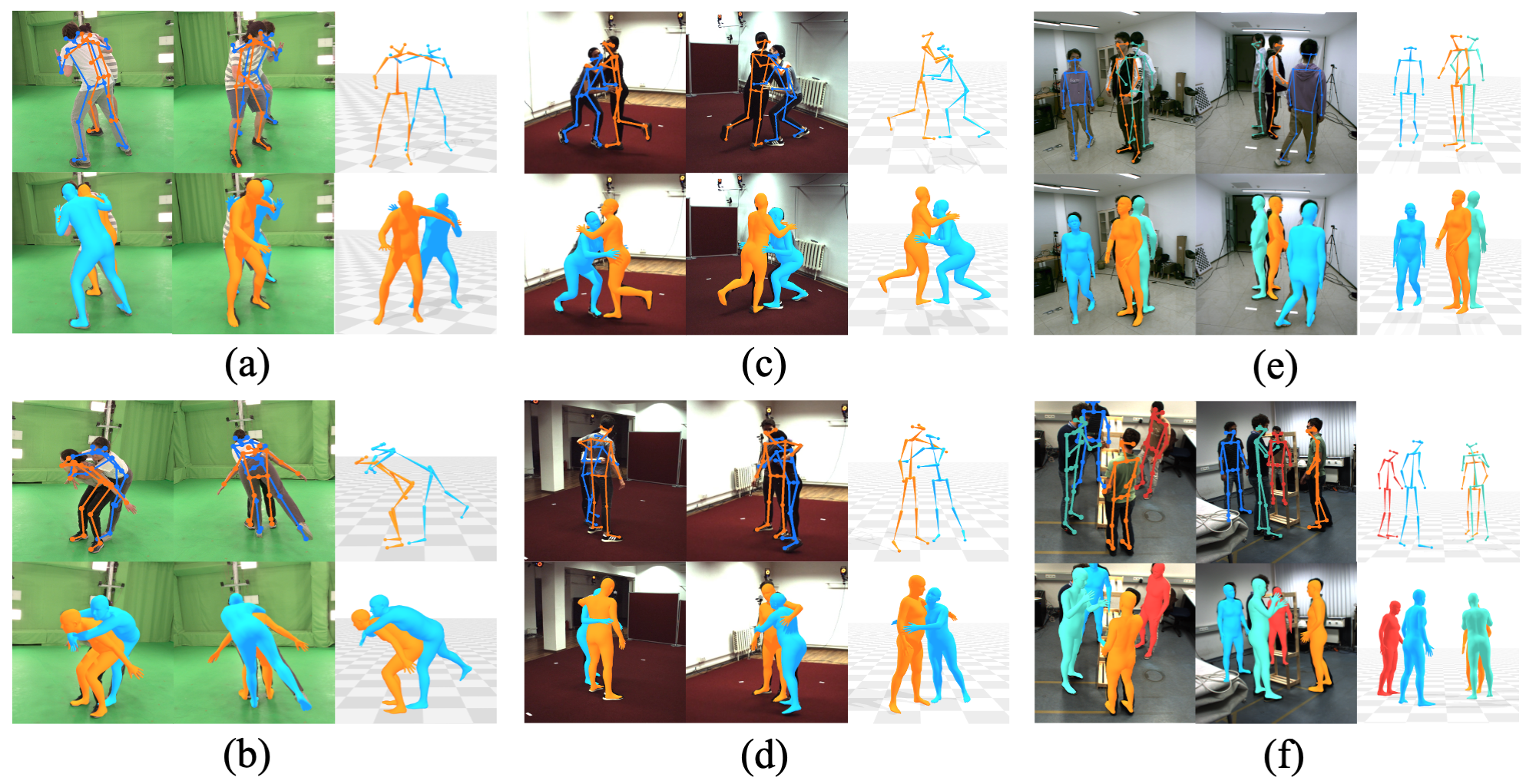}
    \caption{\textbf{Qualitative Results of our method on Hi4D (a)(b), CHI3D (c)(d), MultiHuman Real-Cap (e), and Shelf (f).} The left and middle columns in each sub-figure show the 2D projections of the estimated 3D skeletons and SMPL body meshes on two views. The right column in each sub-figure demonstrates skeletons and SMPL bodies in 3D scenes. }
    \label{fig:demo}
\end{figure*}

\subsection{Additional Qualitative Samples}

\Cref{fig:demo} demonstrates more qualitative results
of our method on Hi4D, CHI3D, and MultiHuman Real-Cap with challenging and close interactions among 2 or 3 people. We also demonstrate results on the Shelf~\cite{shelf} dataset, which contains 4 people without close contact. More results can be found in the Supp Mat.


\subsection{Ablation Study}
To validate the effectiveness of our method, we conduct a detailed analysis of different design choices of our algorithm. All the experiments are conducted on the Hi4D Dataset.

\begin{figure}[h]
  \centering
   \includegraphics[width=0.5\linewidth]{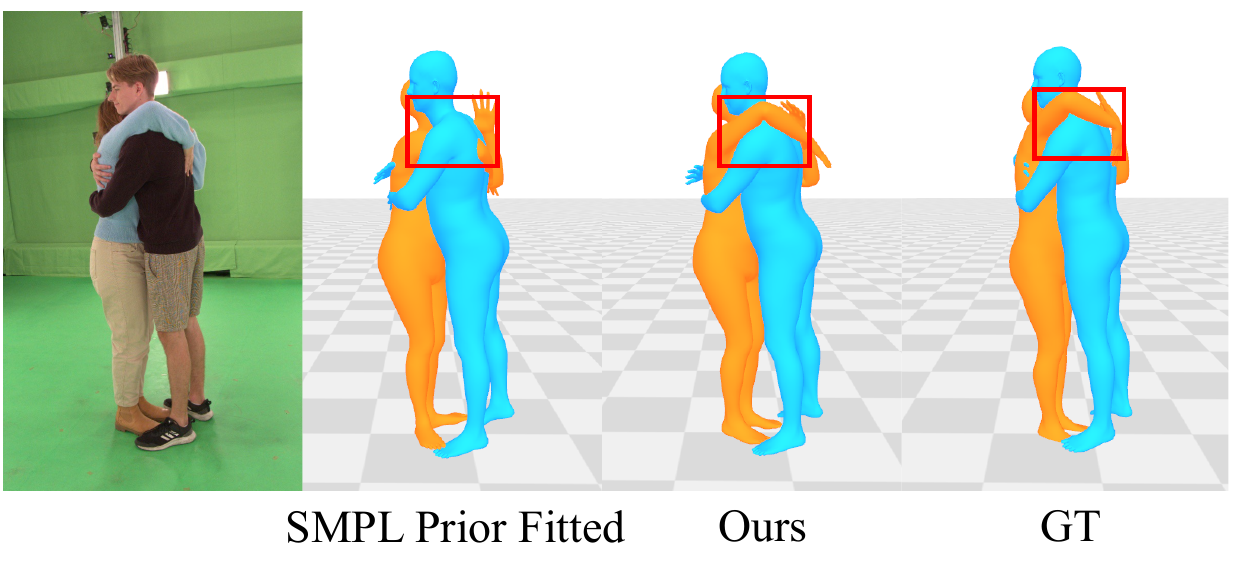}
   \caption{\textbf{ Comparison with SMPL Body Prior.} Only fitting SMPL to 2D observations, some joints in close contact such as arms are incorrectly estimated and even cause intersections between body surfaces. In contrast, our personalized prior enables accurate estimation of poses.}
   \label{fig:ablation_smpl}
\end{figure}

\paragraph{Comparison with SMPL Body Prior.}
To validate the effectiveness of our personalized avatar prior, we compare with a baseline that optimizes SMPL parameters to align 3D joint reprojections to 2D observations. Our method significantly outperforms the SMPL prior baseline (\Cref{tab:ablation}) by leveraging color and silhouette rendering loss to refine poses, reducing reliance on inaccurate joint detections when occlusion happens.  \Cref{fig:ablation_smpl} shows that while the baseline method incorrectly estimates the arm pose and even leads to penetration, our method reconstructs the poses accurately.


\begin{table}[h]
  \centering
  \begin{tabular}{@{}lcccccc@{}}
    \toprule
    \textbf{Method} & \textbf{MPJPE}(mm) $\downarrow$ & \textbf{PCP}(\%) $\uparrow$  &  $\textbf{AP}_{50}$ $\uparrow$ & $\textbf{AP}_{100}$ $\uparrow$ \\
    \midrule
    Ours (SMPL fitted) & 40.41 & 95.10 & 84.04 & 94.37 \\ 
    \midrule
    Ours (w/o RGB loss) & 78.40 &  83.26 & 17.84 & 74.55 \\
    Ours (w/o Mask loss) & 31.00 & 97.33 & 90.28 & 99.06 \\
    \midrule
    Ours (Joint Optimization) & 66.04 & 84.56 & 22.55 & 76.70 \\
    \midrule
    Ours & \textbf{29.37} & \textbf{98.02} & \textbf{96.79} & \textbf{99.06}\\
    \bottomrule
  \end{tabular}
  \caption{\textbf{Quantitative Ablation Results}. Ablations to evaluate our method with only the SMPL fitted method, our method without RGB loss and without Silhouette loss, and our method without alternating optimization. }
  \label{tab:ablation}
\end{table}


\paragraph{Color and Mask Loss.} To demonstrate the effectiveness of the color and silhouette rendering loss in our optimization process, we design baselines without RGB loss and mask loss respectively for comparison. \Cref{tab:ablation} shows that the RGB loss significantly improves pose optimization, and the optimization process will completely deviate from the correct trajectory in the absence of RGB loss. Mask loss is proven to slightly increase the accuracy by adding additional constraints on the rendered human silhouette.

\begin{figure}[h]
  \centering
   \includegraphics[width=0.4\linewidth]{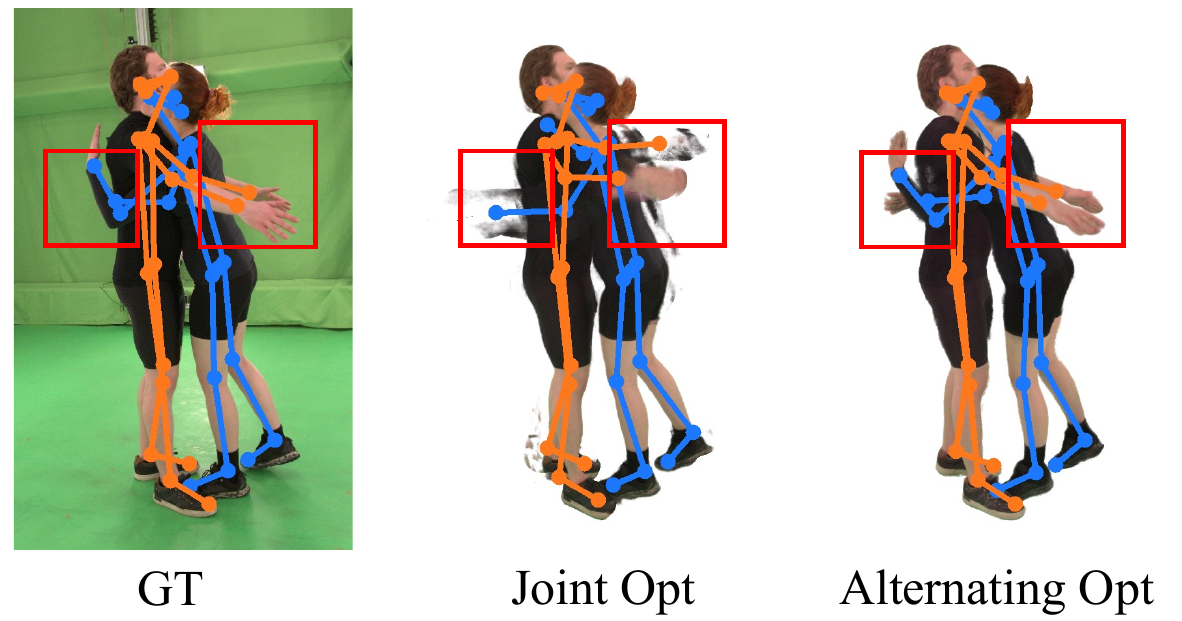}
   \caption{\textbf{Ablation of Alternating Optimization.} We show the results of rendered avatars and projections of the estimated 3D poses. Joint optimization suffers from artifacts around the contact part and in turn causes wrong pose estimations. In contrast, ours reconstructs both avatars and poses correctly. }
   \label{fig:ablation_alter}
\end{figure}

\begin{figure}[h]
  \centering
   \includegraphics[width=0.5\linewidth]{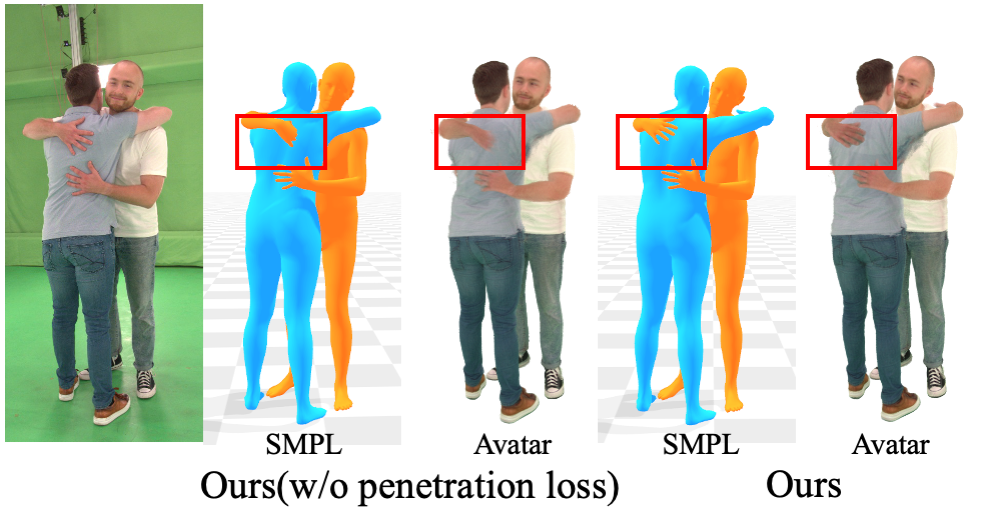}
   \caption{\textbf{Ablation of Penetration Loss.} Without penetration loss, both the avatar and underlying SMPL body tend to have collisions on surfaces.}
   \label{fig:ablation_penetration}
\end{figure}

\paragraph{Alternating Optimization.}
We choose a joint optimization method widely used in avatar reconstructions~\cite{pina, v2a} to compare with our alternating optimization. \Cref{tab:ablation} shows that our method significantly outperforms the chosen baseline. In \Cref{fig:ablation_alter}, the avatars from the baseline suffer from artifacts around contact body parts, leading to wrong 3D pose estimations. In contrast, our method can faithfully reconstruct avatars and poses with challenging initializations.




\paragraph{Penetration Loss.}

Our penetration loss serves an important role in avoiding interpenetration. Comparing our method to the ablated version where we remove this loss, \Cref{fig:ablation_penetration} shows that the SMPL body of one person partially intersects with the other person in the contact area. This is also confirmed by the rendering result of the avatar. By penalizing the collision of density fields of avatars in 3D space, our method largely reduces the penetration and thus achieves more accurate pose estimation.

\subsection{Limitations and Future Work}
Despite greatly improving 98\% of the sequences, our method shows minor improvements when the pose initialization is in the opposite direction or most joints are severely misaligned, leading to local minima in optimization.
Moreover, we do not model hands in our avatar model, it will be a promising direction to integrate hand models~\cite{mano,smplx} into the personalized avatar.  More discussions about limitations and future work can be found in the Supp Mat.
 

\section{Conclusion}
In this paper, we propose AvatarPose, a novel method to estimate the 3D poses of multiple people in close interaction from sparse multi-view videos. Unlike previous methods leveraging SMPL body prior, we leverage the reconstructed avatars as personalized priors to guide pose optimization.  The avatar prior enables us to use color and silhouette observations and to introduce a collision loss in close contact. Our method outperforms SotA methods significantly on public datasets of close human interactions.



\paragraph{Acknowledgements} 
This work was supported by the Swiss SERI Consolidation Grant ”AI-PERCEIVE”. We thank Wenbo Wang, Tianjian Jiang and Juan Zarate for providing us with helpful guidance and feedback.


%
%
\bibliographystyle{splncs04}
\bibliography{main}

\begin{thebibliography}{10}
\providecommand{\url}[1]{\texttt{#1}}
\providecommand{\urlprefix}{URL }
\providecommand{\doi}[1]{https://doi.org/#1}

\bibitem{scape}
Anguelov, D., Srinivasan, P., Koller, D., Thrun, S., Rodgers, J., Davis, J.: Scape: shape completion and animation of people. In: ACM SIGGRAPH 2005 Papers, pp. 408--416 (2005)

\bibitem{belagiannis1}
Belagiannis, V., Amin, S., Andriluka, M., Schiele, B., Navab, N., Ilic, S.: 3d pictorial structures for multiple human pose estimation. In: Proceedings of the IEEE conference on computer vision and pattern recognition. pp. 1669--1676 (2014)

\bibitem{shelf}
Belagiannis, V., Amin, S., Andriluka, M., Schiele, B., Navab, N., Ilic, S.: 3d pictorial structures for multiple human pose estimation. In: Proceedings of the IEEE conference on computer vision and pattern recognition. pp. 1669--1676 (2014)

\bibitem{belagiannis2}
Belagiannis, V., Amin, S., Andriluka, M., Schiele, B., Navab, N., Ilic, S.: 3d pictorial structures revisited: Multiple human pose estimation. IEEE transactions on pattern analysis and machine intelligence  \textbf{38}(10),  1929--1942 (2015)

\bibitem{pandanet}
Benzine, A., Chabot, F., Luvison, B., Pham, Q.C., Achard, C.: Pandanet: Anchor-based single-shot multi-person 3d pose estimation. In: Proceedings of the IEEE/CVF Conference on Computer Vision and Pattern Recognition. pp. 6856--6865 (2020)

\bibitem{ipnet}
Bhatnagar, B.L., Sminchisescu, C., Theobalt, C., Pons-Moll, G.: Combining implicit function learning and parametric models for 3d human reconstruction. In: Computer Vision--ECCV 2020: 16th European Conference, Glasgow, UK, August 23--28, 2020, Proceedings, Part II 16. pp. 311--329. Springer (2020)

\bibitem{smplify}
Bogo, F., Kanazawa, A., Lassner, C., Gehler, P., Romero, J., Black, M.J.: Keep it smpl: Automatic estimation of 3d human pose and shape from a single image. In: Computer Vision--ECCV 2016: 14th European Conference, Amsterdam, The Netherlands, October 11-14, 2016, Proceedings, Part V 14. pp. 561--578. Springer (2016)

\bibitem{dsfn}
Burov, A., Nie{\ss}ner, M., Thies, J.: Dynamic surface function networks for clothed human bodies. In: Proceedings of the IEEE/CVF International Conference on Computer Vision. pp. 10754--10764 (2021)

\bibitem{openpose}
Cao, Z., Simon, T., Wei, S.E., Sheikh, Y.: Realtime multi-person 2d pose estimation using part affinity fields. In: Proceedings of the IEEE conference on computer vision and pattern recognition. pp. 7291--7299 (2017)

\bibitem{animnerf}
Chen, J., Zhang, Y., Kang, D., Zhe, X., Bao, L., Jia, X., Lu, H.: Animatable neural radiance fields from monocular rgb videos. arXiv preprint arXiv:2106.13629  (2021)

\bibitem{dataset_1}
Chen, L., Ai, H., Chen, R., Zhuang, Z., Liu, S.: Cross-view tracking for multi-human 3d pose estimation at over 100 fps. In: Proceedings of the IEEE/CVF conference on computer vision and pattern recognition. pp. 3279--3288 (2020)

\bibitem{snarf}
Chen, X., Zheng, Y., Black, M.J., Hilliges, O., Geiger, A.: Snarf: Differentiable forward skinning for animating non-rigid neural implicit shapes. In: Proceedings of the IEEE/CVF International Conference on Computer Vision. pp. 11594--11604 (2021)

\bibitem{cascaded}
Chen, Y., Wang, Z., Peng, Y., Zhang, Z., Yu, G., Sun, J.: Cascaded pyramid network for multi-person pose estimation. In: Proceedings of the IEEE conference on computer vision and pattern recognition. pp. 7103--7112 (2018)

\bibitem{vtp}
Chen, Y., Gu, R., Huang, O., Jia, G.: Vtp: volumetric transformer for multi-view multi-person 3d pose estimation. Applied Intelligence pp. 1--12 (2023)

\bibitem{higherHRnet}
Cheng, B., Xiao, B., Wang, J., Shi, H., Huang, T.S., Zhang, L.: Higherhrnet: Scale-aware representation learning for bottom-up human pose estimation. In: Proceedings of the IEEE/CVF conference on computer vision and pattern recognition. pp. 5386--5395 (2020)

\bibitem{trackingsam}
Cheng, Y., Li, L., Xu, Y., Li, X., Yang, Z., Wang, W., Yang, Y.: Segment and track anything. arXiv preprint arXiv:2305.06558  (2023)

\bibitem{tempo}
Choudhury, R., Kitani, K.M., Jeni, L.A.: Tempo: Efficient multi-view pose estimation, tracking, and forecasting. In: Proceedings of the IEEE/CVF International Conference on Computer Vision. pp. 14750--14760 (2023)

\bibitem{mvpose2021}
Dong, J., Fang, Q., Jiang, W., Yang, Y., Bao, H., Zhou, X.: Fast and robust multi-person 3d pose estimation and tracking from multiple views. In: T-PAMI (2021)

\bibitem{mvpose}
Dong, J., Jiang, W., Huang, Q., Bao, H., Zhou, X.: Fast and robust multi-person 3d pose estimation from multiple views. In: Proceedings of the IEEE/CVF conference on computer vision and pattern recognition. pp. 7792--7801 (2019)

\bibitem{pina}
Dong, Z., Guo, C., Song, J., Chen, X., Geiger, A., Hilliges, O.: Pina: Learning a personalized implicit neural avatar from a single rgb-d video sequence. In: Proceedings of the IEEE/CVF Conference on Computer Vision and Pattern Recognition. pp. 20470--20480 (2022)

\bibitem{shape_aware}
Dong, Z., Song, J., Chen, X., Guo, C., Hilliges, O.: Shape-aware multi-person pose estimation from multi-view images. In: Proceedings of the IEEE/CVF International Conference on Computer Vision. pp. 11158--11168 (2021)

\bibitem{ershadi}
Ershadi-Nasab, S., Noury, E., Kasaei, S., Sanaei, E.: Multiple human 3d pose estimation from multiview images. Multimedia Tools and Applications  \textbf{77},  15573--15601 (2018)

\bibitem{regress_1}
Fabbri, M., Lanzi, F., Calderara, S., Alletto, S., Cucchiara, R.: Compressed volumetric heatmaps for multi-person 3d pose estimation. In: Proceedings of the IEEE/CVF Conference on Computer Vision and Pattern Recognition. pp. 7204--7213 (2020)

\bibitem{rmpe}
Fang, H.S., Xie, S., Tai, Y.W., Lu, C.: Rmpe: Regional multi-person pose estimation. In: Proceedings of the IEEE international conference on computer vision. pp. 2334--2343 (2017)

\bibitem{chi3d}
Fieraru, M., Zanfir, M., Oneata, E., Popa, A.I., Olaru, V., Sminchisescu, C.: Three-dimensional reconstruction of human interactions. In: Proceedings of the IEEE/CVF Conference on Computer Vision and Pattern Recognition. pp. 7214--7223 (2020)

\bibitem{v2a}
Guo, C., Jiang, T., Chen, X., Song, J., Hilliges, O.: Vid2avatar: 3d avatar reconstruction from videos in the wild via self-supervised scene decomposition. In: Proceedings of the IEEE/CVF Conference on Computer Vision and Pattern Recognition. pp. 12858--12868 (2023)

\bibitem{expi}
Guo, W., Bie, X., Alameda-Pineda, X., Moreno-Noguer, F.: Multi-person extreme motion prediction. In: Proceedings of the IEEE/CVF Conference on Computer Vision and Pattern Recognition. pp. 13053--13064 (2022)

\bibitem{statistical}
Hasler, N., Stoll, C., Sunkel, M., Rosenhahn, B., Seidel, H.P.: A statistical model of human pose and body shape. In: Computer graphics forum. vol.~28, pp. 337--346. Wiley Online Library (2009)

\bibitem{instantavatar}
Jiang, T., Chen, X., Song, J., Hilliges, O.: Instantavatar: Learning avatars from monocular video in 60 seconds. In: Proceedings of the IEEE/CVF Conference on Computer Vision and Pattern Recognition. pp. 16922--16932 (2023)

\bibitem{panoptic}
Joo, H., Liu, H., Tan, L., Gui, L., Nabbe, B., Matthews, I., Kanade, T., Nobuhara, S., Sheikh, Y.: Panoptic studio: A massively multiview system for social motion capture. In: Proceedings of the IEEE International Conference on Computer Vision. pp. 3334--3342 (2015)

\bibitem{mono_1}
Kanazawa, A., Black, M.J., Jacobs, D.W., Malik, J.: End-to-end recovery of human shape and pose. In: Proceedings of the IEEE conference on computer vision and pattern recognition. pp. 7122--7131 (2018)

\bibitem{PARE}
Kocabas, M., Huang, C.H.P., Hilliges, O., Black, M.J.: Pare: Part attention regressor for 3d human body estimation. In: Proceedings of the IEEE/CVF International Conference on Computer Vision. pp. 11127--11137 (2021)

\bibitem{spec}
Kocabas, M., Huang, C.H.P., Tesch, J., M{\"u}ller, L., Hilliges, O., Black, M.J.: Spec: Seeing people in the wild with an estimated camera. In: Proceedings of the IEEE/CVF International Conference on Computer Vision. pp. 11035--11045 (2021)

\bibitem{tokenpose}
Li, Y., Zhang, S., Wang, Z., Yang, S., Yang, W., Xia, S.T., Zhou, E.: Tokenpose: Learning keypoint tokens for human pose estimation. In: Proceedings of the IEEE/CVF International conference on computer vision. pp. 11313--11322 (2021)

\bibitem{plane-sweep}
Lin, J., Lee, G.H.: Multi-view multi-person 3d pose estimation with plane sweep stereo. In: Proceedings of the IEEE/CVF Conference on Computer Vision and Pattern Recognition. pp. 11886--11895 (2021)

\bibitem{seg_motion}
Liu, Y., Gall, J., Stoll, C., Dai, Q., Seidel, H.P., Theobalt, C.: Markerless motion capture of multiple characters using multiview image segmentation. IEEE transactions on pattern analysis and machine intelligence  \textbf{35}(11),  2720--2735 (2013)

\bibitem{smpl}
Loper, M., Mahmood, N., Romero, J., Pons-Moll, G., Black, M.J.: Smpl: A skinned multi-person linear model. ACM transactions on graphics (TOG)  \textbf{34}(6),  1--16 (2015)

\bibitem{single_scene}
Luvizon, D.C., Habermann, M., Golyanik, V., Kortylewski, A., Theobalt, C.: Scene-aware 3d multi-human motion capture from a single camera. In: Computer Graphics Forum. vol.~42, pp. 371--383. Wiley Online Library (2023)

\bibitem{mono_2}
Martinez, J., Hossain, R., Romero, J., Little, J.J.: A simple yet effective baseline for 3d human pose estimation. In: Proceedings of the IEEE international conference on computer vision. pp. 2640--2649 (2017)

\bibitem{mono_3}
Mehta, D., Sridhar, S., Sotnychenko, O., Rhodin, H., Shafiei, M., Seidel, H.P., Xu, W., Casas, D., Theobalt, C.: Vnect: Real-time 3d human pose estimation with a single rgb camera. Acm transactions on graphics (tog)  \textbf{36}(4),  1--14 (2017)

\bibitem{nerf}
Mildenhall, B., Srinivasan, P.P., Tancik, M., Barron, J.T., Ramamoorthi, R., Ng, R.: Nerf: Representing scenes as neural radiance fields for view synthesis. Communications of the ACM  \textbf{65}(1),  99--106 (2021)

\bibitem{selfcontact}
Muller, L., Osman, A.A.A., Tang, S., Huang, C.H.P., Black, M.J.: On self-contact and human pose. In: Proceedings of the IEEE/CVF Conference on Computer Vision and Pattern Recognition (CVPR). pp. 9990--9999 (June 2021)

\bibitem{instantngp}
M{\"u}ller, T., Evans, A., Schied, C., Keller, A.: Instant neural graphics primitives with a multiresolution hash encoding. ACM Transactions on Graphics (ToG)  \textbf{41}(4),  1--15 (2022)

\bibitem{star}
Osman, A.A., Bolkart, T., Black, M.J.: Star: Sparse trained articulated human body regressor. In: Computer Vision--ECCV 2020: 16th European Conference, Glasgow, UK, August 23--28, 2020, Proceedings, Part VI 16. pp. 598--613. Springer (2020)

\bibitem{agora}
Patel, P., Huang, C.H.P., Tesch, J., Hoffmann, D.T., Tripathi, S., Black, M.J.: Agora: Avatars in geography optimized for regression analysis. In: Proceedings of the IEEE/CVF Conference on Computer Vision and Pattern Recognition. pp. 13468--13478 (2021)

\bibitem{smplx}
Pavlakos, G., Choutas, V., Ghorbani, N., Bolkart, T., Osman, A.A., Tzionas, D., Black, M.J.: Expressive body capture: 3d hands, face, and body from a single image. In: Proceedings of the IEEE/CVF conference on computer vision and pattern recognition. pp. 10975--10985 (2019)

\bibitem{rebuttal_1}
Pavlakos, G., Zhu, L., Zhou, X., Daniilidis, K.: Learning to estimate 3d human pose and shape from a single color image. In: Proceedings of the IEEE conference on computer vision and pattern recognition. pp. 459--468 (2018)

\bibitem{neuralbody}
Peng, S., Zhang, Y., Xu, Y., Wang, Q., Shuai, Q., Bao, H., Zhou, X.: Neural body: Implicit neural representations with structured latent codes for novel view synthesis of dynamic humans. In: Proceedings of the IEEE/CVF Conference on Computer Vision and Pattern Recognition. pp. 9054--9063 (2021)

\bibitem{deepcut}
Pishchulin, L., Insafutdinov, E., Tang, S., Andres, B., Andriluka, M., Gehler, P.V., Schiele, B.: Deepcut: Joint subset partition and labeling for multi person pose estimation. In: Proceedings of the IEEE conference on computer vision and pattern recognition. pp. 4929--4937 (2016)

\bibitem{tessetrack}
Reddy, N.D., Guigues, L., Pishchulin, L., Eledath, J., Narasimhan, S.G.: Tessetrack: End-to-end learnable multi-person articulated 3d pose tracking. In: Proceedings of the IEEE/CVF Conference on Computer Vision and Pattern Recognition. pp. 15190--15200 (2021)

\bibitem{mano}
Romero, J., Tzionas, D., Black, M.J.: Embodied hands: Modeling and capturing hands and bodies together. arXiv preprint arXiv:2201.02610  (2022)

\bibitem{scanimate}
Saito, S., Yang, J., Ma, Q., Black, M.J.: Scanimate: Weakly supervised learning of skinned clothed avatar networks. In: Proceedings of the IEEE/CVF Conference on Computer Vision and Pattern Recognition. pp. 2886--2897 (2021)

\bibitem{easymocap}
Shuai, Q., Geng, C., Fang, Q., Peng, S., Shen, W., Zhou, X., Bao, H.: Novel view synthesis of human interactions from sparse multi-view videos. In: ACM SIGGRAPH 2022 Conference Proceedings. SIGGRAPH '22, Association for Computing Machinery, New York, NY, USA (2022). \doi{10.1145/3528233.3530704}, \url{https://doi.org/10.1145/3528233.3530704}

\bibitem{closemocap}
Shuai, Q., Yu, Z., Zhou, Z., Fan, L., Yang, H., Yang, C., Zhou, X.: Reconstructing close human interactions from multiple views. ACM Transactions on Graphics  (dec 2023). \doi{10.1145/3618336}

\bibitem{ROMP}
Sun, Y., Bao, Q., Liu, W., Fu, Y., Black, M.J., Mei, T.: Monocular, one-stage, regression of multiple 3d people. In: Proceedings of the IEEE/CVF international conference on computer vision. pp. 11179--11188 (2021)

\bibitem{BEV}
Sun, Y., Liu, W., Bao, Q., Fu, Y., Mei, T., Black, M.J.: Putting people in their place: Monocular regression of 3d people in depth. In: Proceedings of the IEEE/CVF Conference on Computer Vision and Pattern Recognition. pp. 13243--13252 (2022)

\bibitem{te_neural}
Te, G., Li, X., Li, X., Wang, J., Hu, W., Lu, Y.: Neural capture of animatable 3d human from monocular video. In: European Conference on Computer Vision. pp. 275--291. Springer (2022)

\bibitem{voxelpose}
Tu, H., Wang, C., Zeng, W.: Voxelpose: Towards multi-camera 3d human pose estimation in wild environment. In: Computer Vision--ECCV 2020: 16th European Conference, Glasgow, UK, August 23--28, 2020, Proceedings, Part I 16. pp. 197--212. Springer (2020)

\bibitem{hmor}
Wang, C., Li, J., Liu, W., Qian, C., Lu, C.: Hmor: Hierarchical multi-person ordinal relations for monocular multi-person 3d pose estimation. In: Computer Vision--ECCV 2020: 16th European Conference, Glasgow, UK, August 23--28, 2020, Proceedings, Part III 16. pp. 242--259. Springer (2020)

\bibitem{arah}
Wang, S., Schwarz, K., Geiger, A., Tang, S.: Arah: Animatable volume rendering of articulated human sdfs. In: European conference on computer vision. pp. 1--19. Springer (2022)

\bibitem{direct_regression}
Wang, T., Zhang, J., Cai, Y., Yan, S., Feng, J.: Direct multi-view multi-person 3d pose estimation. In: Ranzato, M., Beygelzimer, A., Dauphin, Y., Liang, P., Vaughan, J.W. (eds.) Advances in Neural Information Processing Systems. vol.~34, pp. 13153--13164. Curran Associates, Inc. (2021), \url{https://proceedings.neurips.cc/paper_files/paper/2021/file/6da9003b743b65f4c0ccd295cc484e57-Paper.pdf}

\bibitem{humannerf}
Weng, C.Y., Curless, B., Srinivasan, P.P., Barron, J.T., Kemelmacher-Shlizerman, I.: Humannerf: Free-viewpoint rendering of moving people from monocular video. In: Proceedings of the IEEE/CVF conference on computer vision and pattern Recognition. pp. 16210--16220 (2022)

\bibitem{graph-based}
Wu, S., Jin, S., Liu, W., Bai, L., Qian, C., Liu, D., Ouyang, W.: Graph-based 3d multi-person pose estimation using multi-view images. In: Proceedings of the IEEE/CVF international conference on computer vision. pp. 11148--11157 (2021)

\bibitem{graph}
Wu, S., Jin, S., Liu, W., Bai, L., Qian, C., Liu, D., Ouyang, W.: Graph-based 3d multi-person pose estimation using multi-view images. In: ICCV (2021)

\bibitem{ghum}
Xu, H., Bazavan, E.G., Zanfir, A., Freeman, W.T., Sukthankar, R., Sminchisescu, C.: Ghum \& ghuml: Generative 3d human shape and articulated pose models. In: Proceedings of the IEEE/CVF Conference on Computer Vision and Pattern Recognition. pp. 6184--6193 (2020)

\bibitem{faster_voxelpose}
Ye, H., Zhu, W., Wang, C., Wu, R., Wang, Y.: Faster voxelpose: Real-time 3d human pose estimation by orthographic projection. In: European Conference on Computer Vision. pp. 142--159. Springer (2022)

\bibitem{Hi4D}
Yin, Y., Guo, C., Kaufmann, M., Zarate, J.J., Song, J., Hilliges, O.: Hi4d: 4d instance segmentation of close human interaction. In: Proceedings of the IEEE/CVF Conference on Computer Vision and Pattern Recognition. pp. 17016--17027 (2023)

\bibitem{mono_4}
Zanfir, A., Marinoiu, E., Sminchisescu, C.: Monocular 3d pose and shape estimation of multiple people in natural scenes-the importance of multiple scene constraints. In: Proceedings of the IEEE Conference on Computer Vision and Pattern Recognition. pp. 2148--2157 (2018)

\bibitem{pymaf}
Zhang, H., Tian, Y., Zhou, X., Ouyang, W., Liu, Y., Wang, L., Sun, Z.: Pymaf: 3d human pose and shape regression with pyramidal mesh alignment feedback loop. In: Proceedings of the IEEE/CVF International Conference on Computer Vision. pp. 11446--11456 (2021)

\bibitem{stnerf}
Zhang, J., Liu, X., Ye, X., Zhao, F., Zhang, Y., Wu, M., Zhang, Y., Xu, L., Yu, J.: Editable free-viewpoint video using a layered neural representation. ACM Transactions on Graphics (TOG)  \textbf{40}(4),  1--18 (2021)

\bibitem{voxeltrack}
Zhang, Y., Wang, C., Wang, X., Liu, W., Zeng, W.: Voxeltrack: Multi-person 3d human pose estimation and tracking in the wild. IEEE Transactions on Pattern Analysis and Machine Intelligence  \textbf{45}(2),  2613--2626 (2022)

\bibitem{4Dassociation}
Zhang, Y., An, L., Yu, T., Li, X., Li, K., Liu, Y.: 4d association graph for realtime multi-person motion capture using multiple video cameras. In: Proceedings of the IEEE/CVF conference on computer vision and pattern recognition. pp. 1324--1333 (2020)

\bibitem{lightweight}
Zhang, Y., Li, Z., An, L., Li, M., Yu, T., Liu, Y.: Lightweight multi-person total motion capture using sparse multi-view cameras. In: Proceedings of the IEEE/CVF International Conference on Computer Vision. pp. 5560--5569 (2021)

\bibitem{deepmulticap}
Zheng, Y., Shao, R., Zhang, Y., Yu, T., Zheng, Z., Dai, Q., Liu, Y.: Deepmulticap: Performance capture of multiple characters using sparse multiview cameras. In: Proceedings of the IEEE/CVF International Conference on Computer Vision. pp. 6239--6249 (2021)

\bibitem{heatmap}
Zhou, H., Hong, C., Han, Y., Huang, P., Zhuang, Y.: Mh pose: 3d human pose estimation based on high-quality heatmap. In: 2021 IEEE International Conference on Big Data (Big Data). pp. 3215--3222. IEEE (2021)

\bibitem{quickpose}
Zhou, Z., Shuai, Q., Wang, Y., Fang, Q., Ji, X., Li, F., Bao, H., Zhou, X.: Quickpose: Real-time multi-view multi-person pose estimation in crowded scenes. In: ACM SIGGRAPH 2022 Conference Proceedings. pp.~1--9 (2022)

\end{thebibliography}
\end{document}